# A model-agnostic approach for generating Saliency Maps to explain inferred decisions of Deep Learning Models

Savvas Karatsiolis [1] and Andreas Kamilaris [1,2]

[1] CYENS Centre of Excellence, Nicosia, Cyprus
[2] University of Twente, Department of Computer Science, Enschede, The Netherlands

**Abstract.** The widespread use of black-box AI models has raised the need for algorithms and methods that explain the decisions made by these models. In recent years, the AI research community is increasingly interested in models' explainability since black-box models take over more and more complicated and challenging tasks. Explainability becomes critical considering the dominance of deep learning techniques for a wide range of applications, including but not limited to computer vision. In the direction of understanding the inference process of deep learning models, many methods that provide human comprehensible evidence for the decisions of AI models have been developed, with the vast majority relying their operation on having access to the internal architecture and parameters of these models (e.g., the weights of neural networks). We propose a model-agnostic method for generating saliency maps that has access only to the output of the model and does not require additional information such as gradients. We use Differential Evolution (DE) to identify which image pixels are the most influential in a model's decision-making process and produce class activation maps (CAMs) whose quality is comparable to the quality of CAMs created with model-specific algorithms. DE-CAM achieves good performance without requiring access to the internal details of the model's architecture at the cost of more computational complexity.

## 1. Introduction

In recent years, Artificial Intelligence (AI) has been undertaking or facilitating many technological applications in almost every field of modern society: healthcare, industrial automation and manufacturing, logistics and retail, policy-making and marketing. AI technologies have such a huge impact on the economy and our societies that AI has been called the "new electricity" [1]. AI is expected to transform every industry and create huge economic value in the following years. Regardless of AI's impressive performance in decision-making and problem-solving, which some years ago seemed very difficult for machines to achieve, the AI models designed for probabilistic inference have highly complex architectures while their internal mechanics are hidden behind an overwhelming number of calculations. Such models are often built with deep learning (DL) methodologies or ensembles based on methods like bagging or boosting. Because of the difficulty to understand the logic behind the decisions made during inference, these models are often referred to as "black-box models" [2], [3], characterized as non-transparent entities that can only be probed to provide output for specific input.

The lack of models' explainability is a compound and significant problem in AI as it is not just related to the absence of supporting evidence for their decisions: model explainability can reveal biases caused by flawed training data and can identify possible ethical issues regarding the model's inference. Model biases constitute a major problem in AI and may render a system useless or societally disapproved because of ethical and practical concerns. For example, an algorithm used in United States hospitals to predict which patients would likely need extra medical care, heavily favored white patients over black patients [4] and it was withdrawn until the problem was fixed. Another example is an algorithm used by Amazon for facilitating personnel hiring and it was greatly favoring men over women [5], [6]. Such discriminative functionality could be identified and resolved before the algorithms reached the production phase if model explainability was studied during the development process. Model explainability is also extremely significant for models that facilitate or fully undertake a medical diagnosis. Any AI decision that influences the life of a patient and her family should be supported by evidence and convincingly explained to the doctors that will adopt the machine's decision or ask/perform further tests. Besides the ethical biases revealed by algorithms that explain the models' decisions, such algorithms can also reveal biases in the training data that prevent the model from learning suitable features. For example, Freitas [7] builds his argument upon the amount of bias that data may introduce to a model, based on a neural network application failure story that is not necessarily real: a neural network was trained to discriminate between friendly and enemy tanks. While its performance was very good on the test set, it had a very bad performance when tested in the field. Later, it was discovered that most enemy tank images were taken on cloudy days while most friendly tank images were taken on sunny days. The model learned to discriminate between sunny and cloudy days instead of focusing on the tank types. A similar situation is described by Ribeiro et al [8], where a classifier discriminating between wolves and husky dogs relied solely on the snowy background to identify a wolf.

The extended use of AI algorithms in many fields raised the European Parliament's concerns regarding privacy rights and the conservation of customers' private information by companies. Specifically, in 2018, a set of new clauses were introduced to the General Data Protection Regulation (GDPR) [9] to ensure the civilians' right to meaningful explanations of the logic involved in automated decision-making. Meanwhile, the High-Level Expert Group on AI published the ethics guidelines for trustworthy AI in 2019 [10]. Explaining the decisions of AI models is crucial for preventing unethical use, avoiding discriminative decision-making, and preventing models' failure to perform as expected because of inappropriate training data. It is also crucial for making a sustainable and beneficial transition to an AI era where machines will take over critical decision making that currently requires a lot of effort by human experts.

In this paper, we focus solely on explaining the decisions of AI models that process images (computer vision DL models). The most popular explanation mean in computer vision is saliency maps (SMs). A saliency map is an image having a width and a height equal to the dimensions of the images used as input to the model. The brightness of each pixel in the SM represents how salient the pixel is: the brighter a pixel is, the more its contribution to the decision of the model.

There is a plethora of algorithms that generate SMs to explain the AI models' decisions. These algorithms are categorized as *model-specific* or *model-agnostic*, depending on whether they need information sourced from the model parameters or not. This information is either in the form of gradients flowing from the output towards a specific layer (that could be the input layer) or the activations of an intermediate model layer or both. The model-specific algorithms require such information to quantify the sensitivity of the model's output to a specific pixel at the input. On the contrary, model-agnostic algorithms do not need any further information besides the scores at the output of the model for some specific input. *Model-agnostic algorithms provide actual black-box explainability because they require no interaction with the model besides inference*.

This paper proposes a model-agnostic method for generating SMs that has access only to the output of the model and does not require additional information such as gradients. We use Differential Evolution (DE) to identify which image pixels are the most influential in a model's decision-making process and we call this method DE-CAM. The rest of the paper describes the DE-CAM technique (Sections 3 and 4) after presenting related work in the field (Section 2).

## 2. Related Work

As mentioned above, the algorithms that generate SMs to explain the AI models' decisions are categorized as model-specific or model-agnostic.

### 2.1 Model-specific Algorithms

The most popular model-specific algorithms for generating SMs are GradCAM [11], GradCAM++ [12], Intergraded Gradient [13] and Full-Gradient [14]. We provide a brief description of each algorithm below.

**GradCAM** uses gradient information from the last convolutional layer of the model to attribute saliency to each input pixel. As convolutional layers retain spatial information, the feature map of the last convolutional layer provides a link between the high-level semantics of the input image and their spatial information. GradCam computes the gradient of a particular class with respect to the activations of the last convolutional layer and then the low-resolution heatmap is calculated by summing the feature maps after being scaled by their respective gradient. The final saliency map (having the same dimensions as the input image) is obtained by upsampling the low-resolution heatmap. GradCAM is a popular choice for computing SAs because it performs decently, requires no hyperparameter tuning, and is fast and simple to implement.

**GradCAM++** is an extension of GradCAM and uses a weighted combination of the positive partial derivatives of the last convolutional layer feature maps with respect to a specific class score. This approach improves GradCAM, especially in the case when multiple objects of the selected class are present in the image, a case that may cause problems for GradCAM because it tends to underestimate object representations that appear fewer times in the feature maps than others.

**Integrated Gradient** uses the integral of gradients with respect to inputs along the path from a given baseline (image consisting of pixels heaving either zero or max value) to input. Integrated Gradient relies on two fundamental axioms: sensitivity and implementation invariance [13]. A problem with this method is that the choice of the baseline (black or white image) may cause performance issues during explainability [15].

**FullGradient** aggregates layer-wise gradient maps multiplied by the bias terms. FullGradient provides attributions to both the input and the neurons of intermediate layers which allows the method to satisfy two key properties: completeness and weak dependence [14].

Kindermans et al [16] and Adebayo et al [17] argue that gradient-based methods are exposed to several problems that render them unreliable, with the saturating gradients problem being the most profound one.

## 2.2 Model-agnostic Algorithms

Model-agnostic algorithms do not suffer from the limitations of model-specific algorithms as mentioned above. The most popular model-agnostic algorithms for generating SMs are LIME [8] and RISE [18].

**LIME** achieves interpretability by training an interpretable surrogate model, such as a linear model. The training points of the surrogate model are comprised of model evaluations of sampled data points around a specified input example. Practically, LIME learns an interpretable model by focusing locally around the prediction. LIME often does not work out of the box, especially for small-resolution images, while its parameters need significant fine-tuning [19].

**RISE** operates solely on the input of the model to assign saliency to the input image pixels. It generates $N$ random masks with values sampled from a Gaussian distribution. These masks are multiplied with the image and the resulting images are fed to the model. Then, a saliency map is obtained by linearly combining the masks with the model predictions corresponding to each masked input. The key idea is that the masked inputs that contain significant pixels for the task get a higher output when used for model prediction. The quality of the results produced by this method is much lower than the quality of the results of gradient-based methods; often the saliency maps produced are not interpretable.

For an extensive survey on various explanation methods, the readers are referred to [19]. In this paper we propose a model-agnostic method called Differential Evolution CAM (DE-CAM) that produces results that are comparable to and in some cases better than the results of gradient-based methods, despite not requiring any information regarding the internal structure of the model. DE-CAM combines the advantages of model-agnostic and model-specific methods at the cost of computational time required to run.

## 3. DE-CAM Method

Our proposed method aims at creating an SM for a certain image by identifying which pixels of the image contribute the most to the identification of a specific class. Instead of exploiting information sourced from the architecture of the model (layer activations, gradients, etc.), DE-CAM solves the optimization problem of finding which set of pixels $p$ in image $X$ of height $H$ and width $W$ maximizes the output of the model for a certain class while satisfying $|p| \ll |W \times H|$. Let $M$ be a 2-dimensional binary mask of height $H$ and width $W$ and its element $m_{i,j}$ at coordinates $(i,j)$ has a binary value, i.e., $m_{i,j} \in \{0,1\}$. Mask $M$ is multiplied with the input image $X$ to provide a modified image $X'$ such as $X' = X \odot M$ that feeds the model producing a new output score for a specific class. Since mask $M$ is 2-dimensional, each element $m_{i,j}$ preserves or eliminates (i.e. by changing to a zero value) all channels of a pixel in image $X$, at the coordinates $(i, j)$. DE-CAM solves the following optimization problem:

$$\underset{M}{argmax} \ f(X \odot M) - \alpha \frac{\sum m_{i,j}}{H \times W}\Big|_{\forall i \in [1,H], \forall j \in [1,W]} \quad (1)$$

where $\alpha$ is the weight of the second term and $f(.)$ represents the model function without the softmax layer (i.e., the logits' values). The second term of the objective function serves the requirement of favoring masks that have a larger number of eliminated pixels: SMs should attribute significance only to the subset of the pixels that contribute the most to the output of a specific class. Since SMs need to be grayscale (and not binary) images, the solution obtained by DE-CAM (optimal binary mask $M$) is not a valid SM. Thus, we apply a Monte Carlo approach for obtaining the SM from a large set of binary masks that are in the vicinity of the optimal solution: DE-CAM computes hundreds of decent solutions to the problem and combines them to obtain the SM. Particularly, the final SM is the result of a process during which each candidate solution assigns a score to the pixels of the SM. Candidate solutions are defined as the individuals of the evolutionary algorithm at convergence time that have a fitness greater than $\frac{2}{3}$ of the mean population fitness.

DE-CAM uses Differential Evolution (DE) [20]–[22] to evolve a population of $N$ candidate solutions to the optimization problem. DE is a good fit for the specific optimization problem because it can deal with real-valued problems, constantly improving the general fitness of the population and pushing the solutions towards search areas that tend to provide high fitness values. This is important when applying the Monte Carlo step to many feasible solutions. We represent each individual (candidate mask) in the population as a set of $K$ ellipses. Each ellipse contains pixels with a value equal to *1* and thus represents which image pixels are preserved after masking the input image. Each ellipse is defined by its left-most point and its right-most point of its bounding box and its rotation such as $[x_0, y_0, x_1, y_1, r]$. Ellipses defined with their top-left and their bottom-right points only (without a rotation angle) are aligned to the *x-y* axes, and they only differ in their position in space and their size. Using a rotational angle in the ellipses' definition provides infinite shape orientations that add the flexibility to construct every possible combination of ellipses (and mask shapes). This allows the algorithm to sustain great mask diversity which is important for eliminating or preserving various configurations of image pixels. Figure 1 shows how a single ellipse is encoded to a "gene" by the evolutionary algorithm.

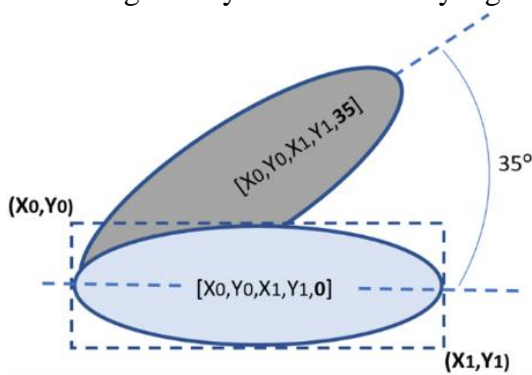

**Figure 1**. The encoding scheme of the evolutionary algorithm: each ellipse of the $K$ ellipses contained in an individual is defined by the top-left/bottom-right coordinates of its containing bounding box and its rotational angle. This definition allows the creation of $K$ ellipses per individual that have different sizes and orientations and their combination produces diverse masks. In this example, the resulting mask spans the area of the two ellipses.

Since each individual consists of $K$ ellipses' definitions, it has a total size of $K \times 5$ genes. The combination of the $K$ ellipses creates the mask that selects which pixels from the image are preserved and which are eliminated. The operation of the DE-CAM is demonstrated in Figure 2 showing how DE-CAM evolves a population of masks to solve the optimization problem defined in Equation (1).

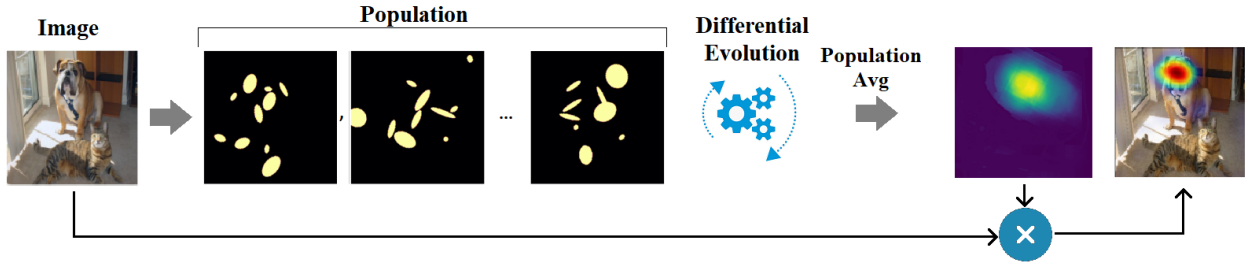

**Figure 2.** Candidate masks with $K = 10$ are manipulated by the evolutionary algorithm of the DE-CAM. Each mask contains $K$ ellipses of various sizes and orientations. In the right-most mask, only 9 ellipses are visible because one ellipse fully overlaps with another. The algorithm evolves a population of masks and aims at computing masks that identify the pixels of the image that contribute to the score of the output class the most while maintaining the number of selected pixels as small as possible.

DE-CAM restrains the size of the ellipses contained in the individuals during the evolutionary process: the minimum distance between the coordinates of the ellipses' bounding boxes is set to $\frac{H}{20}$ and the maximum distance is set to $\frac{H}{4}$, assuming a square input image with $H = W$. Enforcing a minimum distance prevents the use of very small ellipses that are not useful while enforcing a maximum distance prevents the use of large ellipses that can compromise the requirement of maintaining only a fraction of the image pixels that are invaluable to the model for achieving high output on the specific class. The DE-CAM algorithm's pseudocode is shown in Algorithm 1. The values of the DE algorithm's hyperparameters, i.e., the Crossover Probability (CR), the Differential Weight (F) and the number of ellipses in every individual (K) were determined experimentally by applying the DE-CAM with various hyperparameters on different DL models

ALGORITHM 1: DE-CAM

$CR = 0.2$, $F = 0.8$, $K = 10$, $MaxIter = 200$, Fitness $= f(X \odot M) - \alpha \frac{\sum m_{i,j}}{H \times W}$

Initialize a population $P$ of $N$ solutions (individuals). Each solution contains $K$ masks.

Repeat until $MaxIter$ is reached

    Repeat for every individual $x$ in $P$

        Randomly choose 3 different individuals $a, b, c$

        Repeat for each gene (parameter) $i$ in $x$

            With probability $CR$ : $x'_i = a_i + F(b_i - c_i)$ else: $x'_i = x_i$

            Clip the genes (parameters) of $x'$ to comply with min/max size requirements

        If Fitness($x'_i$) > Fitness($x_i$) replace $x_i$ with $x'_i$ in $P$

SumMasks = Sum all $x$ in $P$ that have a Fitness $> \frac{2\,PopMeanFitness}{3}$

SM = SumMasks/MAX(SumMasks)  *#score assignment*

## 4. DE-CAM Evaluation

We tested DE-CAM on ImageNet [23] classification explaining the decisions of three popular models: VGG19 [24], ResNet50 [25] and MobileNetv2 [26]. A visual comparison of our results for the VGG19 model with GradCAM, FullGradient and IntegratedGradient is shown in Figure 3. An analogous comparison is shown for ResNet50 and the MobileNetv2 model in Figures 4 and 5. Considering all approaches listed in related work (see Section 2), we decided to compare DE-CAM with GradCAM, FullGradient and IntegratedGradient, because these are the most popular model-specific algorithms for explaining models' decisions. GradCAM++ could be a good candidate for comparison as well, however, GradCAM++ specializes in multi-object images which was not the focus of this paper (see future work, Section 5).

The results suggest that all methods tend to identify similar areas as being important for the classification, but the computed SMs occasionally exhibit some subtle differences. For example, DE-CAM used on the snail image with VGG19 (Figure 3, last row) attributes more saliency to the shell of the snail and less to its head and tentacles. DE-CAM also attributes importance to the face of the person playing the violin when used on the violin image with ResNet50 (Figure 4, 3$^{rd}$ row) while other methods concentrate solely on the violin. Finally, DE-CAM perceives the uniform of the hockey player as the most important part of the image while other methods also focus on the athlete's stick (Figure 5, 2$^{nd}$ row). In Figure 6 it is evident that the SMs created for Mobilenetv2 by the other methods attribute saliency to larger image regions while DE-CAM's SMs attribute importance to a smaller number of pixels.

To evaluate the results, we employed the Insertion-Deletion metrics [18]. These metrics measure how the confidence of the model changes when we delete or insert certain features. To get the deletion metric we start from the original image and iteratively mute pixels (i.e., we make their value equal to zero) in the order of importance score indicated by the SM. Analogously, to get the insertion metric, we start from a blurred image and start restoring the pixel values in the order of importance score as provided by the SM. An SM's quality is reflected by exhibiting a small Area Under Curve (AUC) for the deletion process and a large AUC for the insertion process, which suggests that the SM conveys reliable information regarding the importance of the image pixels. Figure 6 shows cases of the Insertion-Deletion evaluation for several examples contained in Figures 3-5.

To get a single-value evaluation metric for comparing the performance of each method, we computed the mean difference between $AUC_{insertion}$ and $AUC_{deletion}$ for the SMs of $1K$ images in the ImageNet test set, assuming that the magnitude of $DiffAUC = \Delta(AUC_{insertion}, AUC_{deletion})$ reflects the quality of the computed SMs (we aim at very high $AUC_{insertion}$ and very small $AUC_{deletion}$). We computed this evaluation metric for each model studied and the results are shown in Table 1.

TABLE 1. $DiffAUC$ evaluation on ImageNet test set.

|  | VGG19 | ResNet50 | MobileNet_v2 |
|---|---|---|---|
| GradCAM | 66.5 | 71.8 | 61.2 |
| FullGrad | 60.3 | **73.4** | 64.4 |
| DE-CAM | **66.8** | 70.5 | **64.6** |

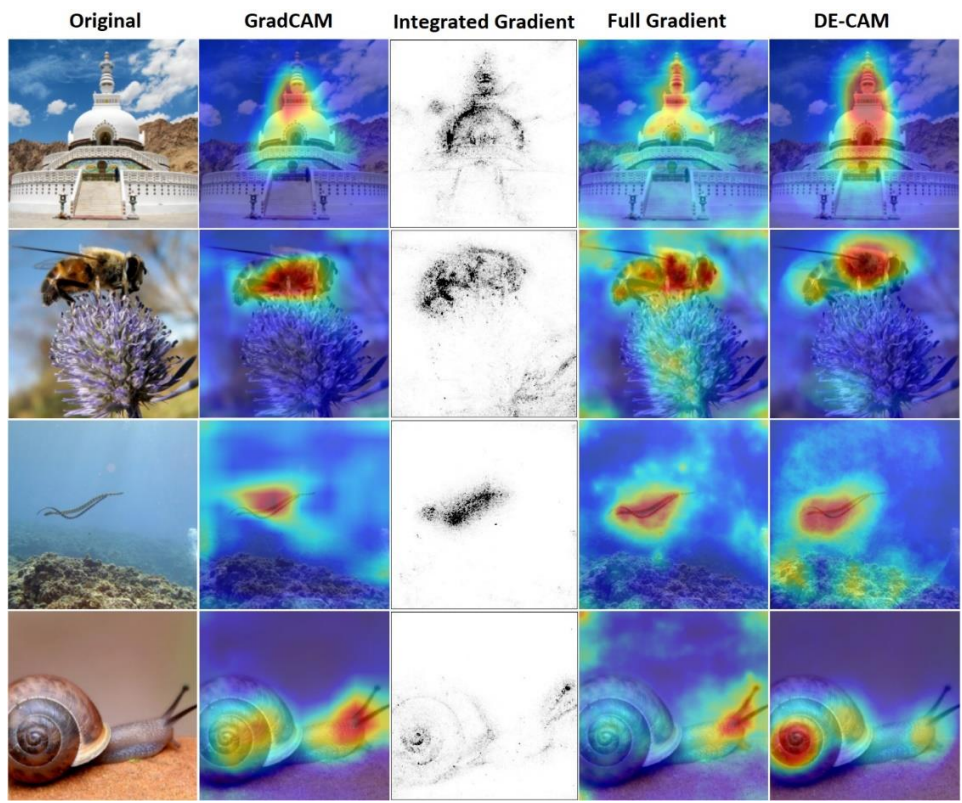

**Figure 3.** SMs produced for the images shown in the left-most column by various methods and DE-CAM with VGG19. The classes of the images from top to bottom are stupa, bee, sea snake and snail.

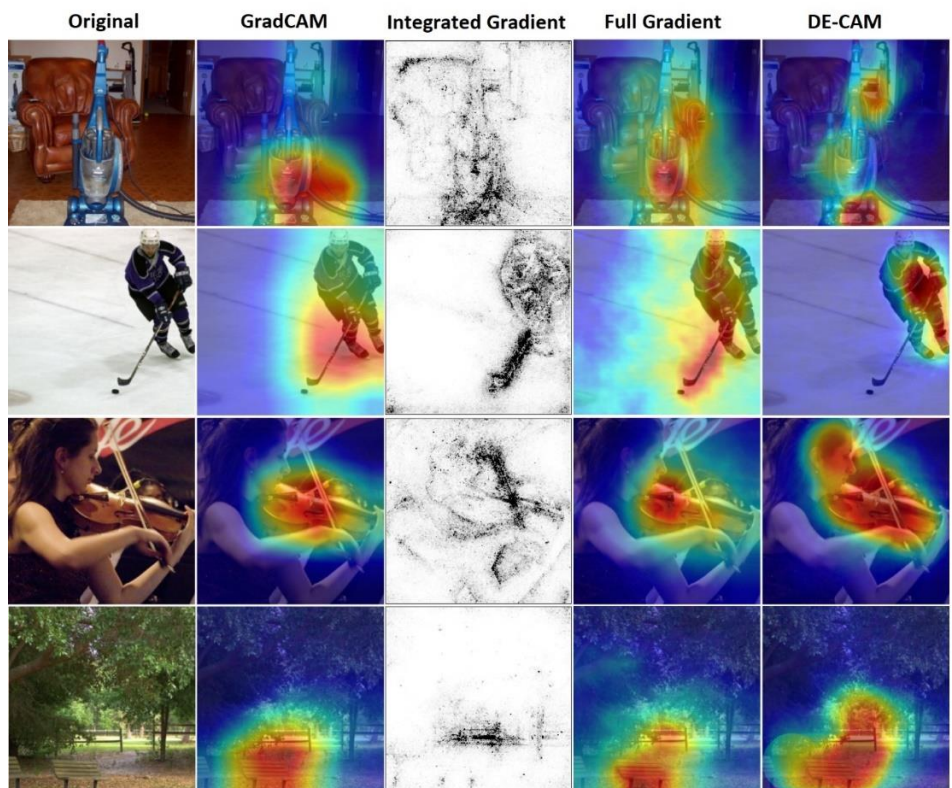

**Figure 4.** SMs produced for the images shown in the left-most column by various methods and DE-CAM for ResNet50. The classes of the images from top to bottom are vacuum, puck, violin and park bench.

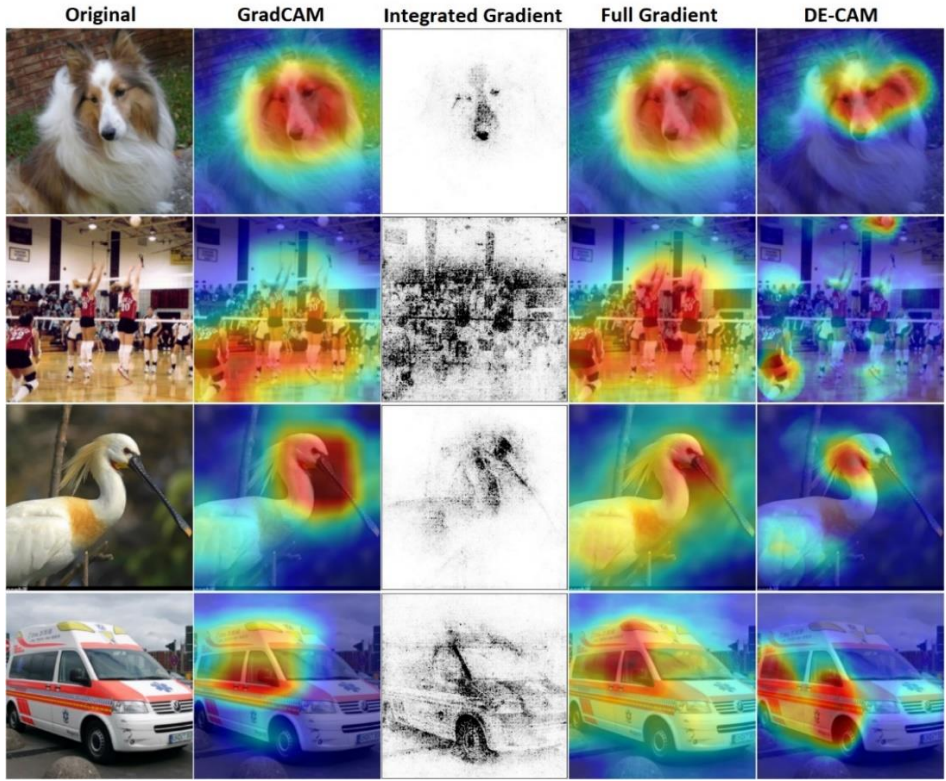

**Figure 5.** SMs produced for the images shown in the left-most column by various methods and DE-CAM for Mobilenet_v2. From top to bottom: Shetland sheepdog, volleyball, spoonbill and ambulance.

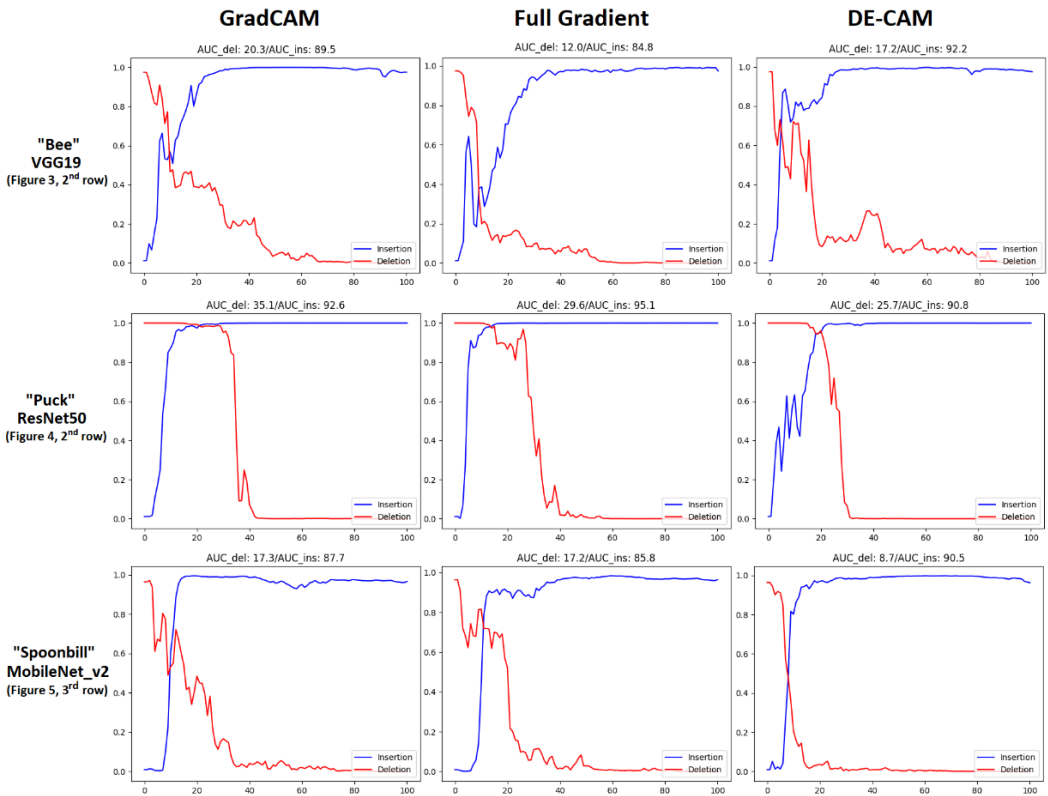

**Figure 6.** Evaluation of SMs shown in Figures 3-5 with the Insertion-Deletion metrics. In each graph, the two curves show how the classification changes by inserting/deleting pixels. The x-axis shows the percentage of the pixels inserted or deleted in the image and the y-axis shows the classification score.

Based on the Insertion-Deletion metrics, DE-CAM surpasses the GradCAM and FullGradient methods on the task of explaining the results of the VGG19 and Mobilenetv2 models. However, it has a lower score than the other two methods when computing SMs for the ResNet50 model. These results are interesting because DE-CAM is a model-agnostic method and does not use the gradients of the model or any information regarding its architecture and still its performance is comparable with the performance of model-specific models.

DE-CAM uses a DE algorithm to search for candidate SMs which requires a lot of iterations and classifications/assessments from the AI model to compute the fitness function of each solution. For example, DE-CAM's computational complexity is orders of magnitude higher than the computational complexity of GradCAM. Being a model-agnostic method, DE-CAM must make a great number of attempts to discover the combinations of pixels that are impactful on the score of a specific class. However, the way the DE algorithm works allows for execution in a multi-processing manner which can reduce the execution time significantly. Furthermore, computing the fitness function of the individuals of the population can be done in large batches and in parallel on multiple Graphical Processing Units (GPUs). For the execution of DE-CAM, we used a multi-processing implementation and two RTX3060 GPUs achieving execution times of *1* minute per run, which is comparable to the execution time of Integrated Gradient and a little higher than the execution time of AblationCAM [27].

## 5. Conclusions

This paper proposes the DE-CAM method, a model-agnostic method for computing SMs for deep learning models based on a differential evolution algorithm. Through various experiments, we showed that DE-CAM computes SMs of similar quality to model-specific methods like GradCAM and FullGradient without using any information about the architecture and the inner mechanisms of the model whose inference is trying to explain. Although DE-CAM has a generally high computational complexity, the use of multi-processing, parallel and large-batch model inferences make it highly practicable for computing SMs in the model-agnostic regime. In future work, we plan to improve DE-CAM to be even more precise in selecting the salient image pixels and introduce mechanisms that enhance the ability of the algorithm to deal with images that contain multiple objects of the same class.

**Funding:** This project received funding from the European Union's Horizon 2020 Research and Innovation Programme under Grant Agreement No. 739578 and the Government of the Republic of Cyprus through the Deputy Ministry of Research, Innovation and Digital Policy.